\documentclass[runningheads]{llncs}
\usepackage{graphicx}
\usepackage{amsmath}
\usepackage{amsfonts}
\usepackage{multirow}
\usepackage{booktabs}
\usepackage{subfig}

\usepackage[colorlinks=true]{hyperref}

\usepackage{array}
\newcolumntype{P}[1]{>{\centering\arraybackslash}p{#1}}

\begin{document}

\title{Self-Supervised Longitudinal Neighbourhood Embedding}

\titlerunning{Self-Supervised Longitudinal Neighbourhood Embedding}
%
\author{
Jiahong Ouyang\inst{1} \and
Qingyu Zhao\inst{1} \and
Ehsan Adeli\inst{1} \and
Edith V Sullivan\inst{1} \and
Adolf Pfefferbaum \inst{1,2} \and
Greg Zaharchuk  \inst{1} \and
Kilian M Pohl\inst{1,2}}
\authorrunning{J. Ouyang et al.}
\institute{Stanford University, Stanford CA 94305, USA \and SRI International, Menlo Park CA, 94025, USA\\}
\maketitle              
\begin{abstract}

Longitudinal MRIs are often used to capture the gradual deterioration of brain structure and function caused by aging or neurological diseases. Analyzing this data via machine learning generally requires a large number of ground-truth labels, which are often missing or expensive to obtain. 
Reducing the need for labels, we propose a self-supervised strategy for representation learning named Longitudinal Neighborhood Embedding (LNE). Motivated by concepts in contrastive learning, LNE explicitly models the similarity between trajectory vectors across different subjects. We do so by building a graph in each training iteration defining neighborhoods in the latent space so that the progression direction of a subject follows the direction of its neighbors. This results in a smooth trajectory field that captures the global morphological change of the brain while maintaining the local continuity.
We apply LNE to longitudinal T1w MRIs of two neuroimaging studies: a dataset composed of 274 healthy subjects, and Alzheimer's Disease Neuroimaging Initiative (ADNI, $N=632$). The visualization of the smooth trajectory vector field and superior performance on downstream tasks demonstrate the strength of the proposed method over existing self-supervised methods in extracting information associated with normal aging and in revealing impact of neurodegenerative disorders. The code is available at \url{https://github.com/ouyangjiahong/longitudinal-neighbourhood-embedding}.

\end{abstract}
\section{Introduction}
Although longitudinal MRIs enable noninvasive tracking of the gradual effect of neurological diseases and environmental influences on the brain over time \cite{whitwell2008longitudinal}, the analysis is complicated by the complex covariance structure characterizing a mixture of time-varying and static effects across visits \cite{garcia2017statistical}. Therefore, training deep learning models on longitudinal data typically requires a large amount of samples with accurate ground-truth labels, which are often expensive or infeasible to acquire for some neuroimaging applications \cite{carass2017longitudinal}.


Recent studies 
suggest that the issue of inadequate labels can be alleviated by self-supervised learning, the aim of which is to automatically learn representations by training on pretext tasks (i.e., tasks that do not require labels) before solving the supervised downstream tasks \cite{kolesnikov2019revisiting}. State-of-the-art self-supervised models are largely based on contrastive learning \cite{oord2018representation,sabokrou2019self,caron2020unsupervised,hassani2020contrastive,tian2019contrastive}, i.e., learning representations by teaching models the difference and similarity of samples. For example, prior studies have generated or identified similar or dissimilar sample pairs (also referred to as positive and negative pairs) based on data augmentation \cite{chen2020simple}, multi-view analysis \cite{tian2019contrastive}, and organizing samples in a lookup dictionary \cite{he2020momentum}. Enforcing such an across-sample relationship in the learning process can then lead to more robust high-level representations for downstream tasks \cite{liu2020self}.


Despite the promising results of contrastive learning on cross-sectional data \cite{li2018non,balakrishnan2018unsupervised,dalca2019unsupervised}, the successful application of these concepts to longitudinal neuroimaging data still remains unclear. In this work, we propose a self-supervised learning model for longitudinal data by exploring the similarity between `trajectories’. Specifically, the longitudinal MRIs of a subject acquired at multiple visits characterize gradual aging and disease progression of the brain over time, which manifests a temporal progression trajectory when projected to the latent space. Subjects with similar brain appearances are likely to exhibit similar aging trajectories. As a result, the trajectories from a cohort should collectively form a smooth trajectory field that characterizes the morphological change of the brain development over time. We hypothesize that regularizing such smoothness in a self-supervised fashion can result in a more informative latent space representation, thereby facilitating further analysis of healthy brain aging and effects of neurodegenerative diseases.

To achieve the smooth trajectory field, we build a dynamic graph in each training iteration to define a neighborhood in the latent space for each subject. The graph then connects nearby subjects and enforces their progression directions
to be maximally aligned (Fig.~\ref{fig:overview}). As such, the resulting latent space captures the global complexity of the progression while maintaining the local continuity of the nearby trajectory vectors. We name the trajectory vectors learned from the neighborhood as \textit{Longitudinal Neighbourhood Embedding (LNE)}.


We evaluate our method on two longitudinal structural MRI datasets: one consists of 274 healthy subjects with the age ranging from 20 to 90, and the second is composed of 632 subjects from ADNI to analyze the progression trajectory of Normal Control (NC), static Mild Cognitive Impairment (sMCI), progressive Mild Cognitive Impairment (pMCI), and Alzheimer's Diease (AD). On these datasets, the visualization of the latent space in a 2D space confirms that the smooth trajectory vector field learned by the proposed method encodes continuous variation with respect to brain aging.
When evaluated on downstream tasks, we obtain higher squared-correlation (R2) in age regression and better balanced accuracy (BACC) in ADNI classifications using our pre-trained model compared to alternative self-supervised or unsupervised pre-trained models.

\section{Method}

We now describe LNE, a method that smooths trajectories in the latent space by local neighborhood embedding. While trajectory regularization has been explored in 2D or 3D spaces (e.g., pedestrian trajectory \cite{alahi2016social,gupta2018social} and in non-rigid registration  \cite{balakrishnan2019voxelmorph}), there are several challenges in the context of longitudinal MRI analysis: (1) each trajectory is measured on sparse and asynchronous (e.g., not aligned by age or visit time) time points; (2) the trajectories live in a high-dimensional space rather than a regular 2D or 3D grid space; (3) the latent representations are defined in a variant latent space that is iteratively updated. To resolve these challenges, we first propose a strategy to train based on pairwise data 
and translate the trajectory-regularization problem to the estimation of a smooth vector field, which is then solved by longitudinal neighbourhood embedding on dynamic graphs.

\begin{figure}[t]
\centering
\includegraphics[width=\textwidth]{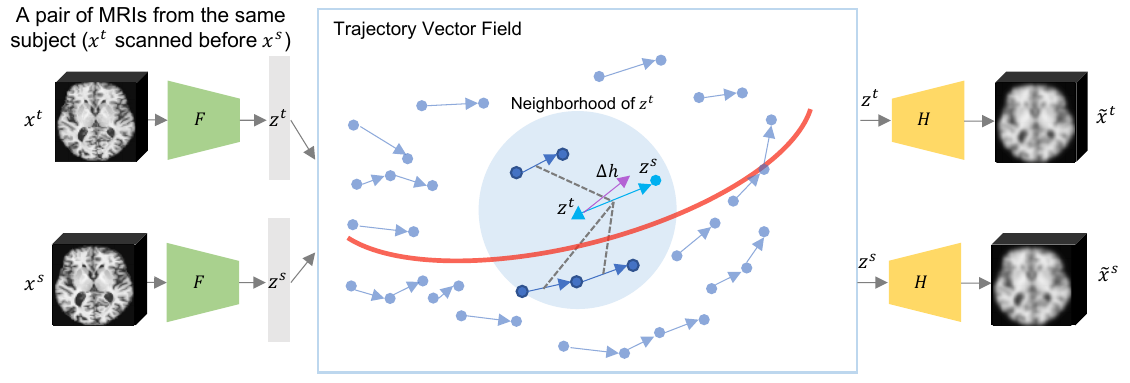}
\vspace{-10pt}
\caption{Overview of the proposed method: 
an encoder projects a subject-specific image pair $(x^t, x^s)$ into the latent space resulting in a trajectory vector (cyan). We encourage the direction of this vector to be consistent with $\Delta h$ (purple), a vector pooled from the neighborhood of $z^t$ (blue circle). As a result, the latent space encodes the global morphological change linked to aging (red curve).
} 
\label{fig:overview}
\end{figure}

\noindent\textbf{Pairwise Training Strategy.} As shown in Fig.~\ref{fig:overview}, each subject is associated with a trajectory (blue vectors) across multiple visits ($\geq 2$) in the latent space. To overcome the problem of the small number of time points in each trajectory, we propose to discretize a trajectory into multiple vectors defined by pairs of images. Compared to using the whole trajectory of sequential images as a training sample as typically done by recurrent neural networks \cite{ouyang2020longitudinal}, this pairwise strategy substantially increases the number of training samples. To formalize this operation, let $\mathcal{X}$ be the collection of all MR images and $\mathcal{S}$ be the set of subject-specific image pairs; i.e., $\mathcal{S}$ contains all $(x^t, x^s)$ that are from the same subject with $x^t$ scanned before $x^s$. These image pairs are then the input to the Encoder-Decoder structure shown in Fig.~\ref{fig:overview}. 
The latent representations generated by the encoder are denoted by $z^t=F(x^t)$, $ z^s=F(x^s)$, where $F$ is the encoder. Then, $\Delta z^{(t,s)} = (z^s - z^t) / \Delta t^{(t,s)}$ is formulated as the normalized trajectory vector, where $\Delta t^{(t,s)}$ is the time interval between the two scans. All $\Delta z^{(t,s)}$ in the cohort define the trajectory vector field. The latent representations are then used to reconstruct the input images by the decoder $H$, i.e., $\tilde{x}^t=H(z^t)$, $ \tilde{x}^s=H(z^s)$. 

\noindent\textbf{Longitudinal Neighbourhood Embedding.} Inspired by social pooling in pedestrian trajectory prediction \cite{alahi2016social,gupta2018social}, we model the similarity between each subject-specific trajectory vector with those from its neighbourhood to enforce the smoothness of the entire vector field. As the high-dimensional latent space cannot be defined by a fixed regular grid (e.g., a 2D image grid space), we propose to define the neighbourhood by building a directed graph $\mathcal{G}$ in each training iteration for the variant latent space that is iteratively updated. The position of each node is defined by the starting point $z^t$ of the vector $\Delta z$ and the value (representation) of that node is $\Delta z$ itself.
For each node $i$, Euclidean distances to other nodes $j\neq i$ are computed by $P_{i,j} = \parallel z^t_i - z^t_j\parallel_2$ while the $N_{nb}$ closest nodes of node $i$ form its 1-hop neighbourhood $\mathcal{N}_i$ with edges connected to $i$.
The adjacency matrix $A$ for $\mathcal{G}$ is then defined as:
\begin{align*} 
    A_{i,j} := 
   \begin{cases}
    exp(-\frac{P_{i,j}^2}{2\sigma_i^2})  & j \in \mathcal{N}_i\\
    0, & j \notin \mathcal{N}_i
    \end{cases}~. \\
    \mbox{with } \sigma_i := max(P_{i,j \in \mathcal{N}_i}) - min(P_{i,j \in \mathcal{N}_i})
    \label{eqn:adj}
\end{align*}

Next, we aim to impose a smoothness regularization on this graph-valued vector field. Motivated by the graph diffusion process \cite{klicpera2019diffusion}, we regularize each node's representation by a \textit{longitudinal neighbourhood embedding} $\Delta h$ `pooled' from the neighbours' representations.
For node $i$, the neighbourhood embedding can be computed by:
\begin{equation*}
\Delta h_i := \sum _{j \in \mathcal{N}_i} A_{i,j} D^{-1}_{i,j} \Delta z_j,
\end{equation*}
where $D$ is the `out-degree matrix' of graph $\mathcal{G}$, a diagonal matrix that describes the sum of the weights for outgoing edges at each node. As shown in Fig.~\ref{fig:overview}, the blue circle illustrates the above operation of learning the neighbourhood embedding that is shown by the purple arrow.



\noindent\textbf{Objective Function.} 
As shown in \cite{zhao2020lssl}, the speed of brain aging is already highly heterogeneous within a healthy population, and subjects with neurodegenerative diseases may exhibit accelerated aging. Therefore, instead of replacing $\Delta z$ with $\Delta h$, we define $\theta_{\langle \Delta z,\Delta h \rangle}$ as the angle between $\Delta z$ and $\Delta h$, and only encourage $\cos(\theta_{\langle \Delta z,\Delta h \rangle}) = 1$, i.e., a zero-angle between the subject-specific trajectory vector and the pooled trajectory vector that represents the local progression direction. As such, it enables the latent representations to model the complexity of the global progression trajectory as well as the consistency of the local trajectory vector field. To impose the direction constraint in the autoencoder, we propose to add this cosine loss for each image pair to the standard mean squared error loss
, i.e.,
\begin{align*}
L := \mathbf{E}_{(x^t, x^s) \sim \mathcal{S}} \left(\parallel x^t - \tilde{x}^t \parallel_2^2 + \parallel x^s - \tilde{x}^s \parallel_2^2 - \lambda \cdot \cos(\theta_{\langle \Delta z,\Delta h \rangle})\right),
\end{align*}
with $\lambda$ being the weighing parameter and $\mathbf{E}$ define the expected value. The objective function encourages the low-dimensional representation of the images to be informative while maintaining a smooth progression trajectory field in the latent space. As the cosine loss is only locally imposed, the global trajectory field can be non-linear, which relaxes the strong assumption in prior studies (e.g., LSSL \cite{zhao2020lssl}) that aging must define a globally linear direction in the latent space. Note, our method can be regarded as a contrastive self-supervised method. For each node, the samples in its neighbourhood serve as positive pairs with the cosine loss being the corresponding contrastive loss.

\section{Experiments}
\noindent\textbf{Dataset.} To show that LNE can successfully disentangle meaningful aging information in the latent space, we first evaluated the proposed method on predicting age from 582 MRIs of 274 healthy individuals with the age ranging from 20 to 90. Each subject had 1 to 13 scans with an average of 2.3 scans spanning an average time interval of 3.8 years. The second data set comprised 2389 longitudinal T1-weighted MRIs (at least two visits per subject) from ADNI, which consisted of 185 NC (age: 75.57 $\pm$ 5.06 years), 119 subjects with AD (age: 75.17 $\pm$ 7.57 years), 193 subjects diagnosed with sMCI (age: 75.63 $\pm$ 6.62 years), and 135 subjects diagnosed with pMCI (age: 75.91 $\pm$ 5.35 years). There was no significant age difference between the NC and AD cohorts (p=0.55, two-sample \textit{t}-test) as well as the sMCI and pMCI cohorts (p=0.75). 
All longitudinal MRIs were preprocessed by a pipeline composed of denoising, bias field correction, skull striping, affine registration to a template, re-scaling to a $64\times64\times64$ volume, and transforming image intensities to z-scores. 

\noindent\textbf{Implementation Details.} Let C$_k$ denote a Convolution(kernel size of $3\times3\times3$)-BatchNorm-LeakyReLU(slope of 0.2)-MaxPool(kernel size of 2) block with $k$ filters, and CD$_k$ an Convolution-BatchNorm-LeakyReLU-Upsample block. The architecture was designed as C$_{16}$-C$_{32}$-C$_{64}$-C$_{16}$-CD$_{64}$-CD$_{32}$-CD$_{16}$-CD$_{16}$ with a convolution layer at the top for reconstruction. The regularization weights were set to $\lambda_{dir}=1.0$ and $\lambda_{recon}=2.0$. The networks were trained for 50 epochs by the Adam optimizer with learning rate of $5 \times 10^{-4}$ and weight decay of $10^{-5}$. To make the algorithm computationally efficient, we built the graph dynamically on the mini-batch of each iteration. A batch size $N_{bs}=64$ and neighbour size $N_{nb}=5$ were used. 

\noindent\textbf{Evaluation.} Five-fold cross-validation (folds split based on subjects) was conducted with 10\% training subjects used for validation. Random flipping of brain hemispheres, and random rotation and shift were used as augmentation during training. We first qualitatively illustrated the trajectory vector field ($\Delta z$) in 2D space by projecting the 1024-dimensional bottleneck representations ($z^t$ and $z^s$) to their first two principal components. We then estimated the global trajectory of the vector field by a curve fitted by robust linear mixed effect model, which considered a quadratic fixed effect with random effect of intercepts. We further quantitatively evaluated the quality of the representations by using them for downstream tasks. On the dataset of healthy subjects, we used the representation $z$ to predict the chronological age of each MRI to show that our latent space was stratified by age. Note, learning a prediction model for normal aging is an emerging approach for understanding structural changes of the human brain and quantifying impact of neurological diseases (e.g. estimating brain age gap \cite{smith2020brain}). R2 and root-mean-square error (RMSE) were used as accuracy metrics. For ADNI, we predicted the diagnosis group associated with each image pair based on both $z$ and trajectory vector $\Delta z$ to highlight the aging speed between visits (an important marker for AD). In addition to classifying NC and AD, we also aimed to distinguish pMCI from sMCI, a significantly more challenging classification task.

The classifier was designed as a multi-layer perceptron containing two fully connected layers of dimension 1024 and 64 with LeakyReLU activation. 
In a separate experiment, we fine-tuned the LNE representation by incorporating the encoder into the classification models. We compared the BACC (accounting for different number of training samples in each cohort) to models using the same architecture with encoders pre-trained by other representation learning methods, including unsupervised methods (AE, VAE \cite{kingma2013auto}), self-supervised method (SimCLR \cite{chen2020simple}), and longitudinal self-supervised method (LSSL \cite{zhao2020lssl}). 

\begin{figure}[t!]
\centering
\includegraphics[width=0.95\linewidth]{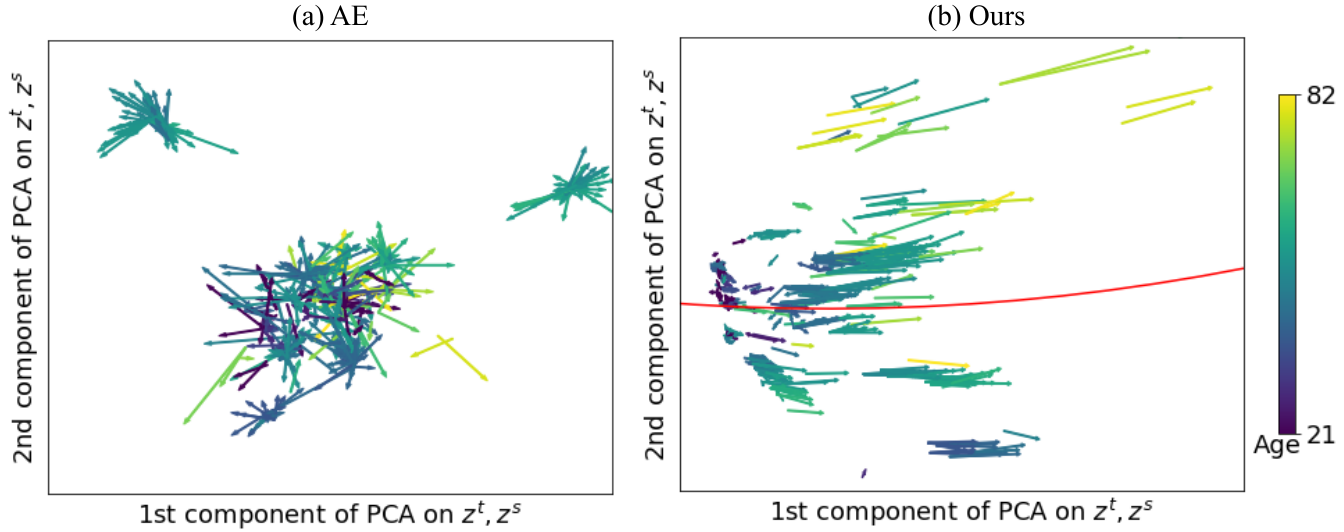}%
\vspace{-7pt}
\caption{Experiments on healthy aging: Latent space of AutoEncoder (AE) (a) and the proposed LNE (b) projected into 2D PCA space of $z^t$ and $z^s$. Arrows represent $\Delta z$ and are color-coded by the age of $z^t$. The global trajectory in (b) is fitted by robust linear mixed effect model (red curve).}
\label{fig:lab}
\end{figure}

\subsection{Healthy Aging}
Fig.~\ref{fig:lab} illustrates the trajectory vector field derived on one of the 5 folds by the proposed method (Fig.~\ref{fig:lab}(b)). We observe LNE resulted in a smooth vector field that was in line with the fitted global trajectory shown by the red curve in Fig.~\ref{fig:lab}(b). Moreover, chronological age associated with the vectors (indicated by the color) gradually increased along the global trajectory (red curve), indicating the successful disentanglement of the aging effect in the latent space. Note, such continuous variation in the whole age range from 20 to 90 was solely learned by self-supervised training on image pairs with an average age interval of 3.8 years (without using their age information). Interestingly, the length of the vectors tended to increase along the global trajectory, suggesting a faster aging speed for older subjects. On the contrary, without regularizing the longitudinal changes, AE did not lead to clear disentanglement of brain age in the space (Fig.~\ref{fig:lab}(a)). 

As shown in Table \ref{tab:res} (left), we utilized the latent representation $z$ to predict the chronological age of the subject. In the scenario that froze the encoder, the proposed method achieved the best performance with an R2 of 0.62, which was significantly better ($p<0.01$, t-test on absolute errors) than the second-best method LSSL with an R2 of 0.59. In addition to R2, the RMSE metrics are given in the supplement Table S1, which also suggests that LNE achieved the most accurate prediction. These results align with the expectation that a pre-trained self-supervised model with explicitly modeling of aging effect can lead to better downstream supervised age prediction. Lastly, when we fine-tuned the encoder during training, LNE remained as the most accurate method (both LNE and LSSL achieved an R2 of 0.74).

\begin{figure}[t!]
\centering
\includegraphics[width=0.95\linewidth]{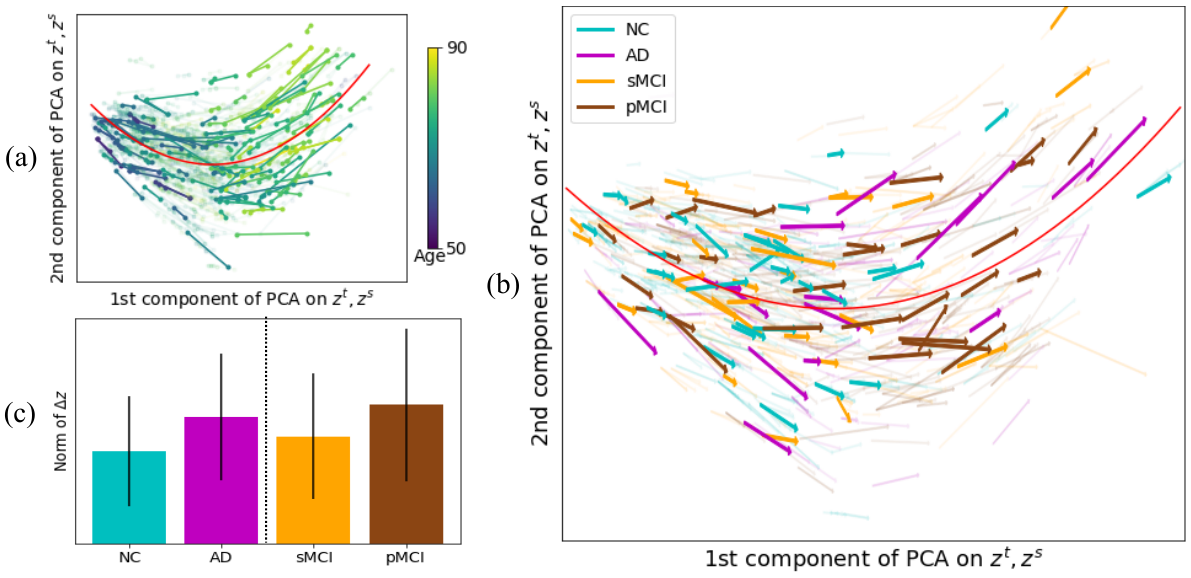}%
\vspace{-7pt}
\caption{Experiments on ADNI: (a) The age distribution of the latent space. Lines connecting $z^t$ and $z^s$ are color-coded by the age of $z^t$; 
Red curve is the global trajectory fitted by a robust linear mixed effect model. (b) Trajectory vector field color-coded by diagnosis groups; (c) The norm of $\Delta z$ encoding the speed of aging for 4 different diagnosis groups.}
\label{fig:adni}
\end{figure}

\subsection{Progression of Alzheimer's Disease}
We also evaluated the proposed method on the ADNI dataset. All 4 cohorts (NC, sMCI, pMCI, AD) were included in the training of LNE as the method was impartial to diagnosis groups (did not use labels for training). Similar to the results of the prior experiment, the age distribution in the latent space in Fig.~\ref{fig:adni}(a) suggests a continuous variation with respect to brain development along the global trajectory shown by the red curve. We further illustrated the trajectory vector field by diagnosis groups in Fig.~\ref{fig:adni}(b). While the starting points ($z^t$) of different diagnosis groups mixed uniformly in the field, vectors of AD (pink) and pMCI (brown) were longer than NC (cyan) and sMCI (orange). This suggests that LNE stratified the cohorts by their `speed of aging' rather than age itself, highlighting the importance of using longitudinal data for analyzing AD and pMCI. This observation was also evident in Fig.~\ref{fig:adni}(c), where AD and pMCI had  statistically larger norm of $\Delta z$ than the other two cohorts (both with $p<0.01$). This finding aligned with previous AD studies \cite{toepper2017dissociating} suggesting that AD group has accelerated aging effect compared to the NC group, and so does the pMCI group compared to the sMCI group. 

The quantitative results on the downstream supervised classification tasks are shown in Table \ref{tab:res} (right). As the length of $\Delta z$ was shown to be informative, we concatenated $z^t$ with $\Delta z$ as the feature for classification (classification accuracy based on $z^t$ only is reported in the supplement Table S2). The representations learned by the proposed method yielded significantly more accurate predictions than all baselines ($p<0.01$, DeLong's test). Note that the accuracy of our model with the frozen encoder even closely matched up to other methods after fine-tuning. This was to be expected because only our method and LSSL explicitly modeled the longitudinal effects which led to more informative $\Delta z$. In addition, our method that focused on local smoothness could capture the potentially non-linear effects underlying the morphological change along time, while the `global linearity' assumption in LSSL may lead to information loss in the representations. It is worth mentioning that reliably distinguishing the subjects that will eventually develop AD (pMCI) from other MCI subjects (sMCI) is crucial for timely treatment. To this end, Supplement Table S3 suggests LNE improved over prior studies in classifying sMCI vs. pMCI, highlighting potential clinical values of our method.


\begin{table}[!t]
\centering
\small
\begin{tabular}{c|P{1.5cm}|P{1.5cm}||P{1.5cm}|P{1.5cm}|P{1.5cm}|P{1.5cm}}
    \toprule
    \multirow{3}{*}{Methods} & \multicolumn{2}{c||}{Health Aging \textbf{(R2)}} & \multicolumn{4}{c}{ADNI \textbf{(BACC)}} \\
    \cline{2-7}
     & \multicolumn{2}{c||}{Age} & \multicolumn{2}{c|}{NC vs AD} & \multicolumn{2}{c}{sMCI vs pMCI}  \\
    \cline{2-7}
     & Frozen & Fine-tune & Frozen & Fine-tune & Frozen & Fine-tune\\
    \hline
    No pretrain & - & 0.72 & - & 79.4 & - & 69.3 \\
    AE & 0.53 & 0.69 & 72.2 & 80.7 & 62.6 & 69.5 \\
    VAE \cite{kingma2013auto} & 0.51 & 0.69 & 66.7 & 77.0 & 61.3 & 63.8 \\
    SimCLR \cite{chen2020simple} & 0.56 & 0.73 & 72.9 & 82.4 & 63.3 & 69.5 \\
    LSSL \cite{zhao2020lssl} & 0.59 & \textbf{0.74} & 74.2 & 82.1 & 69.4 & 71.2 \\
    Ours (LNE) & \textbf{0.62} & \textbf{0.74}  & \textbf{81.9} & \textbf{83.6} & \textbf{70.6} & \textbf{73.4}\\
 \bottomrule
\end{tabular}
\caption{Supervised downstream tasks in frozen or fine-tune scenarios. Left: Age regression on healthy subjects with R2 as an evaluation metric. Right: classification on ADNI dataset with BACC as the metric.} 
\label{tab:res}
\end{table}

\section{Conclusion}
In this work, we proposed a self-supervised representation learning framework, called LNE, that incorporates advantages from the repeated measures design in longitudinal neuroimaging studies. By building the dynamic graph and learning longitudinal neighbourhood embedding, LNE yielded a smooth trajectory vector field in the latent space, while maintaining a globally consistent progression trajectory that modeled the morphological change of the cohort. It successfully modeled the aging effect on healthy subjects, and enabled better chronological age prediction compared to other self-supervised methods. Although LNE was trained without the use of diagnosis labels, it demonstrated capability of differentiating diagnosis groups on the ADNI dataset based on the informative trajectory vector field. When evaluated for downstream task of classification, it showed superior quantitative classification performance as well. 

\noindent{\bf Acknowledgement:} This work was supported by NIH funding R01 MH113406, AA017347, AA010723, and AA005965.

%
%
\bibliographystyle{splncs04}
\bibliography{mybibliography}

\setcounter{table}{0}
\renewcommand{\thetable}{S\arabic{table}}

\section*{Supplementary}

\begin{table}[!h]
\centering
\small
\begin{tabular}{c|P{1.5cm}|P{1.5cm}|P{1.5cm}|P{1.5cm}}
    \toprule
    \multirow{3}{*}{Methods}  & \multicolumn{4}{c}{Chronological Age Prediction} \\
    \cline{2-5}
     & \multicolumn{2}{c|}{Frozen} & \multicolumn{2}{c}{Fine-tuned}  \\
    \cline{2-5}
     & R2 & RMSE & R2 & RMSE\\
    \hline
    No pretrain & - & - & 0.72 & 8.7  \\
    AE & 0.53 & 11.4 & 0.69 & 9.3 \\
    VAE \cite{kingma2013auto} & 0.51 & 11.6 & 0.69 & 9.4\\
    SimCLR \cite{chen2020simple} & 0.56 & 11.1 & 0.73 & 8.9\\
    LSSL \cite{zhao2020lssl} & 0.59 & 10.8 & \textbf{0.74} & \textbf{8.4} \\
    Ours (LNE) & \textbf{0.62} & \textbf{10.3} & \textbf{0.74} & 8.5\\
 \bottomrule
\end{tabular}
\vspace{5pt}
\caption{Complete results for chronological age prediction on the dataset of 274 healthy subjects (an extension of Table 1(left) of the main text).}
\label{tab:res1}
\end{table}

\begin{table}[!h]
\centering
\small
\begin{tabular}{c|P{2cm}|P{2cm}|P{2cm}|P{2cm}}
    \toprule
    Methods  & \multicolumn{2}{c|}{Classification on NC vs. AD} & \multicolumn{2}{c}{Classification on sMCI vs. pMCI} \\
    \cline{2-5}
    BACC & Frozen & Fine-tuned & Frozen & Fine-tuned\\
    \hline
    No pretrain &  - & 79.9 & - & 64.2 \\
    AE &  70.8 & 80.2 & 60.9 & 65.3  \\
    VAE \cite{kingma2013auto} & 64.3 & 76.7 & 60.8 & 63.4  \\
    SimCLR \cite{chen2020simple} & 72.8 & \textbf{82.6} & \textbf{61.4} & 66.7 \\
    LSSL \cite{zhao2020lssl} & 71.2 & 82.4 & 59.3 & 68.2 \\
    Ours (LNE) & \textbf{74.4} & 82.2 & 59.8 & \textbf{69.6} \\
 \bottomrule
\end{tabular}
\vspace{5pt}
\caption{Results for diagnosis type classification on ADNI using $z$ solely as features (a comparison with Table 1 (right) of the main text using $z + \Delta z$). For sMCI vs. pMCI classification, the proposed method resulted in the most accurate prediction only in the fine-tuning scenario. It suggests that the difference between sMCI and pMCI might be mainly the `progressiveness' of neurodegeneration, which could not be fully encoded
by the single time point's representation and thereby justifies the need for modeling longitudinal effects.}
\label{tab:res2}
\end{table}

\begin{table}[!h]
\centering
\begin{tabular}{cP{1.5cm}P{3cm}P{2cm}P{1.5cm}}
\hline
Method & Type & Modalities & sMCI/pMCI & BACC\\
\hline
\multicolumn{4}{l}{Cross-sectional}\\
Liu et al. \cite{liu2018landmark} & D & MRI & 465/205 & 62.2 \\
Zu et al. \cite{zu2016label} & N & MRI, PET & 56/43 & 69.0 \\
Suk et al. \cite{suk2014hierarchical} & N+D & MRI & 128/76 & 63.8 \\
Lin et al. \cite{lin2018convolutional} & D & MRI & 100/164 & 73.0 \\
Huang et al. \cite{huang2019diagnosis} & D & MRI, PET & 441/326 & 76.9 \\
Zhou et al.(a) \cite{zhou2019latent} & N & MRI, PET, SNP & 205/157 & 74.3 \\
Zhou et al.(b) \cite{zhou2019deep} & N & MRI, PET & 114/71 & 78.3 \\
\hline
\multicolumn{4}{l}{Longitudinal}\\
Gray et al.\cite{gray2012multi} & N & MRI, PET & 64/53 & 62.7 \\
Cui et al.\cite{cui2019rnn} & D & MRI & 236/167 & 71.7 \\
Platero et al.\cite{platero2020predicting} & N & MRI, neuro-psychological tests & 215/206 & 77.1 \\
Ours (LNE) & D & MRI & 193/135 & 73.4 \\
\hline
\end{tabular} 
\vspace{5pt}
\caption{Comparison of the proposed method with other traditional methods and deep-learning-based methods in sMCI/pMCI classification on ADNI dataset. `D' denotes deep-learning methods, and `N' denotes non-deep-learning methods. The proposed method achieved best performance among all methods that were solely based on structural MRI.}
\label{tab:res3}
\end{table}

\end{document}